\documentclass{article}

\usepackage[numbers]{natbib}
\usepackage[preprint]{neurips_2024}
\usepackage{algorithm}
\usepackage{algpseudocode}
\usepackage{hyperref}
\usepackage[utf8]{inputenc} % allow utf-8 input
\usepackage[T1]{fontenc}    % use 8-bit T1 fonts
\usepackage{hyperref}       % hyperlinks
\usepackage{url}            % simple URL typesetting
\usepackage{booktabs}       % professional-quality tables
\usepackage{amsfonts}       % blackboard math symbols
\usepackage{nicefrac}       % compact symbols for 1/2, etc.
\usepackage{microtype}      % microtypography
\usepackage{xcolor}         % colors
\usepackage{makecell}
\usepackage{pifont}
\usepackage{graphicx}
\usepackage{subfigure}
\usepackage{listings}
\usepackage{multirow} 
\usepackage{tcolorbox}
\usepackage{colortbl}
\usepackage{times}
\usepackage{markdown}
\usepackage{latexsym}
\definecolor{lightblue}{RGB}{212,220,247}
\definecolor{orange}{HTML}{FF9F2E}
\definecolor{darkpink}{HTML}{FFB8B6}
\definecolor{darkblue}{HTML}{5B9BD5}
\definecolor{lightred}{RGB}{241,208,207}
\definecolor{darkgreen}{RGB}{50,100,0}

\usepackage{longtable}

\usepackage[utf8]{inputenc}
\usepackage[T1]{fontenc}
\tcbuselibrary{listings, breakable, skins}
\usepackage{listings}

\title{Z1: Efficient Test-time Scaling with Code}

\author{%
  Zhaojian Yu$^{1}$,
  Yinghao Wu$^{1}$,
  Yilun Zhao$^{2}$,
  Arman Cohan$^{2}$,
  Xiao-Ping Zhang$^{1}$
  \\
$^1$Tsinghua University
$^2$Yale University
\\
}

\begin{document}

\maketitle

\begin{abstract}

Large Language Models (LLMs) can achieve enhanced complex problem-solving through test-time computing scaling, yet this often entails longer contexts and numerous reasoning token costs. 
In this paper, we propose an efficient test-time scaling method that trains LLMs on code-related reasoning trajectories, facilitating their reduction of excess thinking tokens while maintaining performance.
First, we create Z1-Code-Reasoning-107K, a curated dataset of simple and complex coding problems paired with their short and long solution trajectories. 
Second, we present a novel Shifted Thinking Window to mitigate overthinking overhead by removing context-delimiting tags (e.g., <think>…</think>) and capping reasoning tokens. 
Trained with long and short trajectory data and equipped with Shifted Thinking Window, our model, Z1-7B, demonstrates the ability to adjust its reasoning level as the complexity of problems and exhibits efficient test-time scaling across different reasoning tasks that matches R1-Distill-Qwen-7B performance with about 30\% of its average thinking tokens.
Notably, fine-tuned with only code trajectories, Z1-7B demonstrates generalization to broader reasoning tasks (47.5\% on GPQA Diamond). 
Our analysis of efficient reasoning elicitation also provides valuable insights for future research.\footnote{Our model, data, and code are open-source at \url{https://github.com/efficientscaling/Z1}}

% First, we curate a reasoning dataset containing code, Z1-Code-Reasoning-107K, which includes simple-to-complex problems evolved over multiple rounds, along with their corresponding short and long solution trajectories.
% Second, we propose a shifted thinking window to reduce the computational overhead caused by overthinking during reasoning. This is achieved by eliminating delimiters (e.g., <think>…</think>) that segment the context window and by terminating model output when the maximum thinking token limit is exceeded. Through training with a combination of long and short trajectory dataset and the shifted thinking window, our Z1-7B model demonstrates efficient test-time scaling, outperforming previous open-source 7B models. Furthermore, Z1, trained solely on code-related trajectories, exhibits improved performance on the GPQA Diamond benchmark, highlighting the generalization of code-related trajectory fine-tuning to other reasoning tasks. We conduct detailed analysis of efficient test-time scaling with code, providing supportive insights for future research.

\end{abstract}

\begin{figure}[h]
    \centering
    \includegraphics[width=\linewidth]{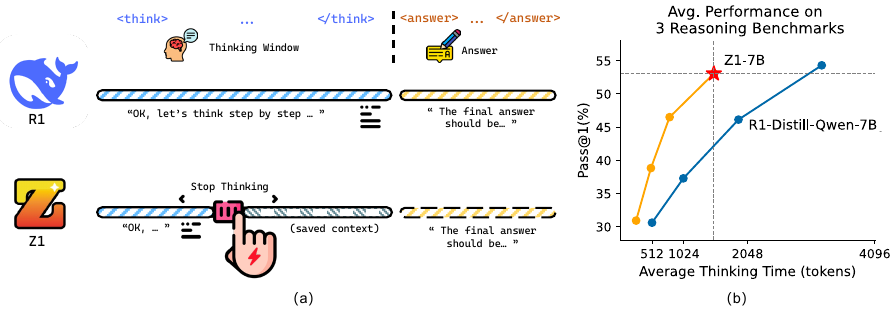} 
    \caption{Comparing Z1 with R1-like models on context window (a) and test-time scaling (b). Z1 models exhibit more efficient test-time compute scaling than R1-Distill-7B, with its shifted thinking window. Z1-7B exhibits efficient test-time scaling across 3 different reasoning tasks (LiveCodeBench, MATH500, GPQA Diamond) and matches R1-Distill-Qwen-7B performance with about 30\% of its average thinking tokens.}
    \label{fig:teaser}
\end{figure}

% \newpage
\section{Introduction}
Large Reasoning Models (LRMs), such as OpenAI o1 ~\cite{jaech2024openai} and DeepSeek R1~\cite{guo2025deepseek}, have demonstrated remarkable advances in complex reasoning tasks through test-time compute scaling~\cite{wei2022chain}, particularly in competitive mathematics and programming. 
These models, trained with large-scale Reinforcement Learning (RL)~\cite{lightman2023let,feng2023alphazero}, have emerged step-by-step reasoning abilities to solve complex problems effectively. However, the elaborate reasoning process also leads to super long contexts and numerous thinking tokens, challenging the efficient utilization of LRMs.

Existing open-source works, such as s1~\cite{muennighoff2025s1} and LIMO~\cite{ye2025limo}, train non-reasoning models into reasoning models with manually curated problems and distilled long chain-of-thought (CoT)~\cite{wei2022chain} trajectories yet do not address the challenges posed by test-time compute scaling with respect to long contexts and an excessive thinking tokens. 
For example, s1~\cite{muennighoff2025s1} introduces budget forcing, which either appends ending words to truncate reasoning processes or extrapolation words (e.g., "wait") to inspire the model to continue thinking, thereby precisely controlling context length. 
While extrapolation words can control model reasoning, direct truncations may disrupt the model's thinking process, consequently degrading performance. 
Thus, we propose the problem: "Is there an efficient test-time scaling way to reduce the model’s thinking tokens consumption while preserving its reasoning performance?".

In this paper, we implement efficient test-time scaling with code-related trajectory fine-tuning and present the shifted thinking mode of LRMs: \textbf{weak reasoning to simple problems, strong reasoning to complex problems}, which significantly reduces the thinking tokens consumption of LRMs in problem solving.
Specifically, we create the Z1-Code-Reasoning-107K dataset, comprising 107K simple and complex code-related problems paired with their reasoning traces distilled from the QwQ-32B-Preview~\cite{qwq32b} model. 
We train Qwen2.5-Coder-7B-Instruct~\cite{hui2024qwen25coder} into a reasoning model, with this 107K code-related long and short trajectory dataset.
We eliminate the context split with delimiters (e.g., <think>...</think>) and introduce novel Shifted Thinking Window: (I) For simple problems, the models generate solutions with in a low reasoning token computation. 
(II) For complex problems, we cap the thinking tokens; if the model outputs exceeds this threshold, we append a hint phrase to the end of reasoning trace, forcing it to produce an answer based on the existing thought process. 
Shifted Thinking Window enables model that get trained with long and short trajectories to adjust their reasoning level as the complexity of problems, thereby avoiding the overthinking of LRMs.
Fine-tuned on long and short reasoning trajectories and equipped with the Shifted Thinking Window, our model \textbf{Z1-7B} exhibits efficient test-time scaling across different reasoning tasks and matches R1-Distill-Qwen-7B performance with about 30\% (as shown in ~\autoref{fig:teaser}) its average thinking tokens.

Furthermore, we conduct data ablation experiments to identify the critical factors driving reasoning elicitation. We design three greedy dataset sampling strategies and train Qwen2.5-Coder-7B-Instruct with these subsets, highlighting two crucial factors (Mean Trajectory Length and Training Sample Size) in trajectory dataset: 
(1) longest-greedy sampling, which prioritizes the longest token-length samples to ensure the inclusion of the most extended reasoning traces; 
(2) shortest greedy sampling, which selects the shortest token-length samples; and 
(3) random sampling under different sample sizes (16K, 64K).
Our results reveal that the model trained on the longest-greedy subset exhibits the best performance, underscoring the importance of trajectory length in training dataset for efficient test-time scaling. 
We also investigate model performance across varying training sample sizes, by comparing the subsets (16K, 64K, full)  of different sizes.
Experimental results demonstrate that the full Z1-Code-Reasoning-107K dataset outperforms all smaller subsets, highlighting the significance of dataset size for reasoning model fine-tuning.

In summary, our contributions are threefold:  
(1) We implement efficient test-time scaling with code and successfully generalize it to reasoning tasks beyond code.  
(2) We propose the shifted thinking window, a method that prevents overthinking by adapting to the complexity of reasoning tasks, significantly improving thinking efficiency of reasoning models.  
(3) Through data ablation studies, we dissect reasoning datasets with code and identify key factors for effective reasoning elicitation.

\section{Z1: Efficient Test-time Scaling with Code}
\begin{figure}[!t]
    \centering
    \includegraphics[width=\linewidth]{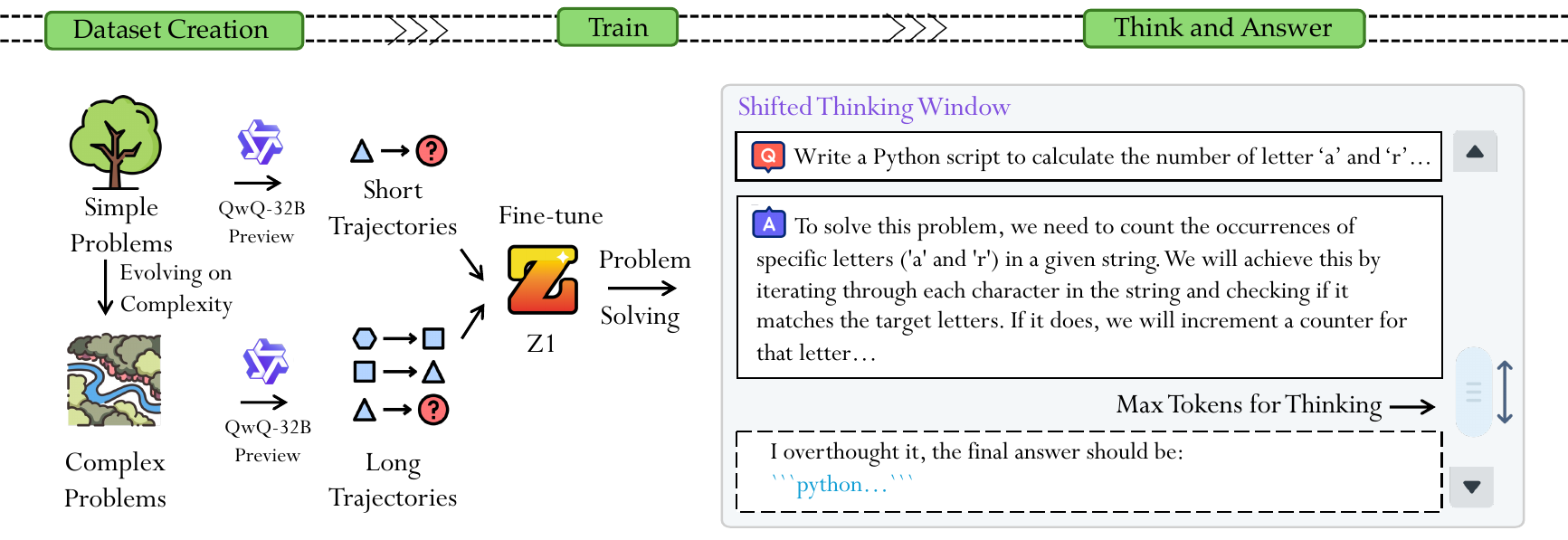} 
    \caption{Overview of Z1 training and inference. Fine-tuned with long and short trajectory data, Z1 could solve simple and complex problems in shifted thinking window efficiently.}
    \label{fig:overview}
\end{figure}

% \begin{figure}[t]
%     \centering
%     \includegraphics[width=0.5\linewidth]{fig/length_comp.pdf}
%     \caption{Caption}
%     \label{fig:enter-label}
% \end{figure}
As illustrated in \autoref{fig:overview}, to achieve efficient test-time scaling, we train non-reasoning models of code-related reasoning trajectories with varying lengths and introduce shifted thinking window to replace the context split with delimiters (e.g., <think>...</think>). After supervised fine-tuning (SFT), our model could adjust its reasoning level according to the complexity of the input problem. In this section, we detail our methodology, including the creation of the reasoning dataset with code (Section \ref{sec:data}) and the implementation of the shifted thinking window (Section \ref{sec:shift}).

\subsection{Dataset Creation}
\label{sec:data}
We create an efficient test-time scaling dataset that integrates both short reasoning trajectories for simple problems and strong reasoning trajectories for complex problems, emphasizing diverse reasoning trajectory lengths in the training set. 
However, existing reasoning trajectory datasets predominantly feature complex problems with long chains of thought (CoT), posing a challenge for training efficient reasoning models due to the lack of short and straightforward trajectories.
To address this, we approach the problem from the perspective of evolving question complexity and select problems from Code Evol-Instruct dataset~\cite{xu2024wizardlm},
% ~\footnote{\url{https://huggingface.co/datasets/theblackcat102/evol-codealpaca-v1}}
which evolves in depth and breadth to cover a wide range of complexities and has proven effective in non-reasoning models~\cite{luo2023wizardcoder}. 
\begin{figure}[b]
    \centering
    \includegraphics[width=\linewidth]{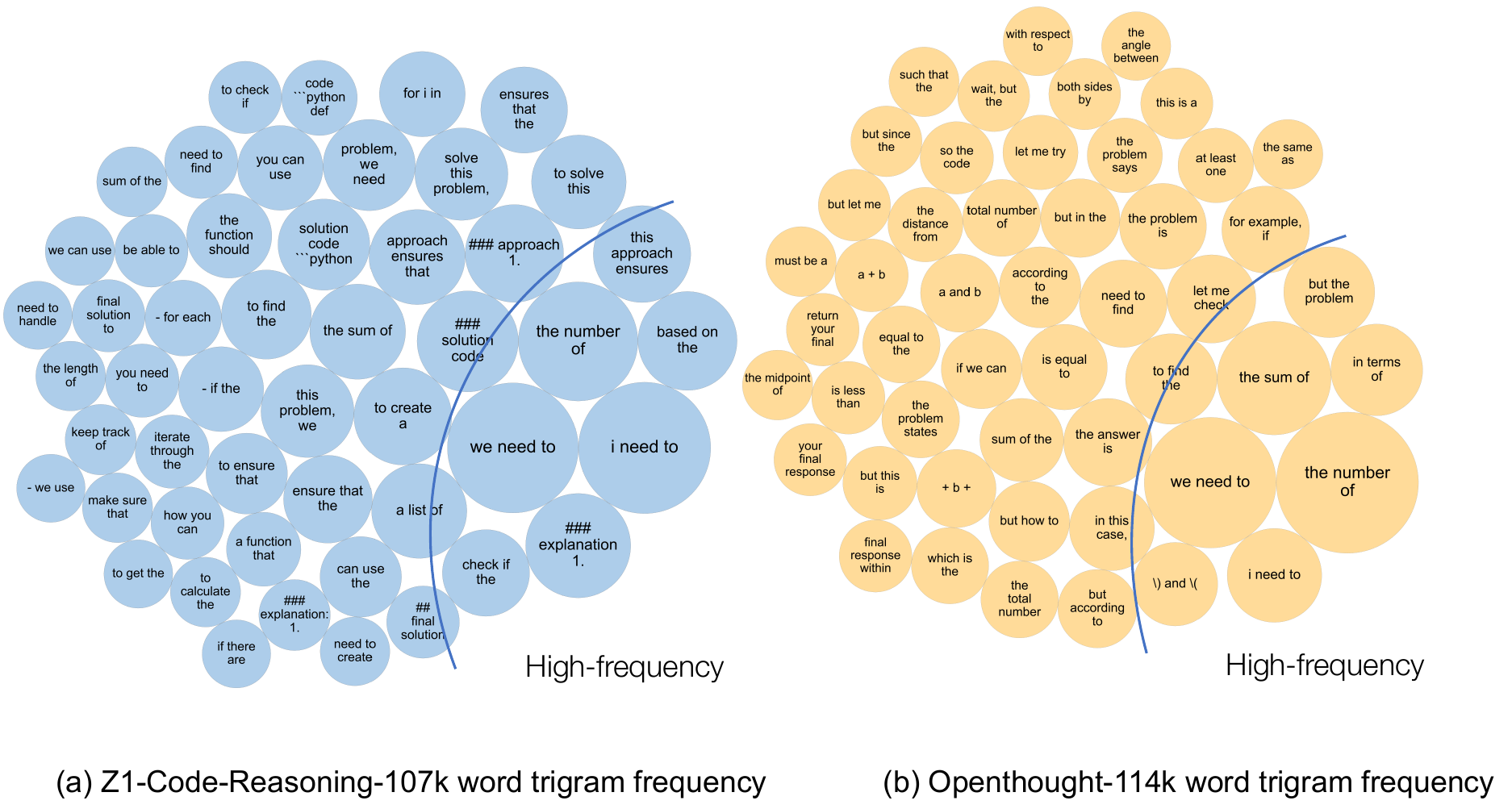} 
    \caption{The comparison between Z1-Code-Reasoning-107K and OpenThoughts-114K. We computed the top-50 most frequent trigrams in both datasets. Circle size reflects word frequency, with larger circles indicating higher frequencies.}
    \label{fig:word_freq}
\end{figure}
We generate reasoning trajectories using the QwQ-32B-preview model and truncated the trajectories length to 8192 tokens, removing approximately 3\% of samples with repetitive reasoning processes to mitigate excessive cyclic thinking in the training data. 
The remaining reasoning trajectories, paired with their problems, constitute the Z1-code-reasoning-107K dataset, with less than 1\% of the data being truncated.

To further analyze our dataset, we compare the top-50 trigram word frequencies of Z1-Code-Reasoning-107K with  OpenThoughts-114K~\cite{openthoughts} dataset\footnote{\url{https://huggingface.co/datasets/open-thoughts/OpenThoughts-114k}}. Z1-code-reasoning-107K exclusively contains code-related reasoning trajectories, while OpenThoughts-114K is a reasoning dataset distilled from DeepSeek R1, featuring 114K high-quality examples spanning math, science, code, and puzzles. \autoref{fig:word_freq} illustrates a trend in the word frequency distributions of Z1-code-reasoning-107K and OpenThoughts-114K: high-frequency trigrams exhibit homogeneity, while mid-frequency trigrams show differentiation. For example, the high-frequency trigrams in both datasets (e.g., "I need to," "we need to," "the number of") indicate the model’s summarization of the next reasoning step, which highlights the commonality between code trajectory and other complex problems.
In contrast, mid-frequency trigrams in Z1-code-reasoning-107K, such as "iterate through the" and "for each," capture loop-based logic characteristic of code-related trajectories, distinct from the mathematical logic exemplified by trigrams (e.g., "a+b", "equal to the") in OpenThoughts-114K. This underscores the unique reasoning patterns inherent in code-related trajectories.

\subsection{Shifted Thinking Window}
\label{sec:shift}
To enforce a "think-before-answer" pattern, existing LRMs like DeepSeek R1 typically use delimiters (e.g., <think>...</think>) to split the context window into two parts, where the model first reasons in the thinking window and then outputs the final answer in the answering window. 
However, this pattern often introduces unnecessary reasoning when processing simple problems that do not require deep thought. 
In the training and inference of our model, we eliminate this context split, allowing the model to flexibly fine-tune and generalize across short and long trajectories and avoiding the overthoughts for simple problems. 
We refer it to Shifted Thinking Window, where the model’s context window is not rigidly divided into two parts by delimiters but instead of a shifted window:
(I) For simple problems, the model fine-tuned on both short and long trajectories can directly output concise reasoning and answers within the context. 
(II) For complex problems, we cap the maximum thinking length within which the model can either reason or provide an answer; if the reasoning trajectory exceeds this maximum length, the end of the model’s output will be appended with a hint phrase to enforce a direct answer.
The essence of shifted thinking: weak reasoning for simple problems and strong reasoning for complex problems significantly reduces unnecessary reasoning by model, thereby demonstrating more efficient test-time compute scalings.

\section{Experiments}

\subsection{Implementation Details}
Following the previous work~\cite{muennighoff2025s1}, we take a model that has already been pretrained and instruction tuned and further finetune it for reasoning. Specifically, we select
Qwen2.5-Coder-Instruct series models, which have already achieves a good performance on various code-related benchmarks. For all training samples, we avoid using delimiters (e.g., <think>...</think>) to separate the whole trajectory into thinking part and the answering part. This adjustment allows the LRM to avoid mandated overthinking, enabling more automatic and efficient test-time scaling: weak reasoning for simple problems and strong reasoning for complex problems. 

We perform supervised finetuning on Qwen-2.5-Coder-7B-Instruct \cite{hui2024qwen25coder} using our Z1-Code-Reasoning-107K dataset, yielding Z1-7B. We do not compute loss on questions, only on reasoning trajectories and solutions. For fine-tuning hyperparameters, we train our model with a learning rate of 1e-5 warmed up linearly for 100 steps and then decayed over the rest of training (836 steps in total training) following a cosine schedule. We train all the models in bfloat16 precision with Pytorch Fully Shard Data Parallel (FSDP) and set a global batch size to 128 for 2 epochs using 8 NVIDIA A100-80G GPUs. In addition, all other settings not mentioned in this paper follow the default values of Huggingface Trainer\footnote{\url{https://huggingface.co/docs/transformers/main_classes/trainer}}.

\subsection{Evaluation Setup}
\paragraph{Benchmarks}
We select three representative reasoning benchmarks covering different topics: 
\textbf{LiveCodeBench}~\cite{jain2024livecodebench} continuously collects new problems over time from contests across three competition platforms, including LeetCode, AtCoder, and CodeForces. 
Unless otherwise specified, we benchmarks LLMs on such competition-level programming tasks with the latest full set (880 problems until Feb, 2025) of LiveCodeBench v5.
\textbf{GPQA Diamond}~\cite{rein2024gpqa} consists of 198 PhD-level science questions from Biology, Chemistry and Physics. Experts with PhDs in the corresponding domains only achieved 69.7\% on GPQA Diamond, which show its inherent difficulty and challenges.
\textbf{MATH500}~\cite{hendrycks2021measuring} is a benchmark of competition math problems of varying difficulty. Following previous work~\cite{guo2025deepseek}, we evaluate our model on the same subset selected by OpenAI~\cite{lightman2023let}.
Alongside the three common reasoning benchmarks, we also incorporate a non-reasoning benchmark:
\textbf{BigCodeBench} is a benchmark mainly focusing on more challenging and practical code generation with complex instructions and diverse function calls. In this section, we adopt the BigCodeBench-Hard-Instruct (148 problems included) subset to evaluate LRM on short trajectory thinking. For all benchmarks, we generate a sample for each question with a temperature of 0 (greedy) to measure accuracy.
Through these benchmarks, we can evaluate the reasoning ability of LLMs from difference perspectives.

\paragraph{Baselines}

We benchmark Z1 against a series of top-tier models: OpenAI o1-series models ~\cite{jaech2024openai}: o1-mini and o1-preview, representing close-source test-time scaling models; Deepseek-R1 series~\cite{guo2025deepseek}: Deepseek-R1, R1-Distill-Qwen (32B and 7B) and Qwen's QwQ-32B-Preview~\cite{qwq-32b-preview}, open-weight reasoning models; Sky-T1-32B-Preview~\cite{sky_t1_2025}, s1.1-7B~\cite{muennighoff2025s1}, OpenThinker-7B~\cite{openthoughts}, open models with open reasoning data; Deepseek-V3~\cite{liu2024deepseek}, GPT-4o~\cite{gpt4o},  Qwen2.5-Coder-7B-Instruct~\cite{hui2024qwen25coder}, four representative non-reasoning models. Our model, Z1, is fully open including weights, reasoning data, and code.
We evaluate Z1 using shifted thinking window with a maximum thinking tokens of 4,096.
For all baseline models, we use the reported results whenever available. 
If no reported scores are provided, we evaluate the model using budget forcing with the configuration provided.

\subsection{Main Result}
\begin{table}[t]

\small
\caption{Results on 4 benchmarks. We evaluate Z1 models with shifted thinking window. For other models without a reported score, we budget force it by adding "the final answer is:".}
\label{tab1:main}
\resizebox{\linewidth}{!}{%
\begin{tabular}{l|c|cccc|c}
\toprule
\textbf{Model}   & {\makecell{\textbf{Data} \\ \textbf{Source}}} & {\makecell{\textbf{MATH} \\ \textbf{500}}}   & {\makecell{\textbf{GPQA} \\ \textbf{Diamond}}} &{\makecell{\textbf{LiveCode} \\ \textbf{Bench}}}  & {\makecell{\textbf{BigCode} \\ \textbf{Bench-Hard}}} & \textbf{AVG}\\
\midrule
\multicolumn{7}{c}{API only}                                                                            \\
\midrule
o1-preview               & N/A & 85.5     & 73.3         & 43.2          & 23.0     &    56.3  \\
o1-mini                  & N/A & 90.0     & 60.0         & 53.7          & 27.7     &   57.9  \\

\midrule
\multicolumn{7}{c}{Open Weights}                                                                        \\
\midrule
Deepseek-R1              & N/A & 97.3     & 71.5         & 77.9          & 29.7     &   67.6    \\
R1-Distill-Qwen-32B & R1/800K & 94.3     & 62.1         & -          & 23.6     &    -   \\
R1-Distill-Qwen-7B  & R1/800K & 83.3     & 49.1         & 40.5          & 3.4      &   44.1   \\
QwQ-32B-Preview          & N/A & 90.6     & 60.0         & 59.9          & 25.0     &    58.9   \\
\midrule
\multicolumn{7}{c}{Non-reasoning Model}                                                               \\
\midrule
Deepseek-V3              & N/A & 90.2     & 59.1         & 56.3          & 27.7     &    58.3   \\
GPT-4o-0513                   & N/A & 75.8     & 46.5         & 43.4          & 25.0     &  47.7  \\
% Codestral-2501          &  & -         & -                & 35.3  & -       &  \\
Qwen2.5-Coder-7B-Ins &    N/A   & 68.6       & 37.4         & 32.3             & 20.3 & 39.7  \\ 

\midrule
\multicolumn{7}{c}{Open Weights and Data}                                                               \\
\midrule

Sky-T1-32B-Preview       & QwQ/17K & 88.6     & 56.8         & -             & 26.4     &   -   \\
s1.1-7B       & R1/1K & 79.2     & 31.8         & 15.2             & 4.7       &     31.7  \\
OpenThinker-7B           & R1/114K & 83.0     & 42.4         & 25.3             & 17.6        &  42.1   \\
 
\midrule

Z1-7B           & QwQ/107K    & 76.4       & 47.5         & 35.3             & 22.3   &  45.4 \\
% Z1-14B           & QwQ/107K    & 80.8       & -         & -             & -   &  - \\  
\bottomrule
\end{tabular}
}
\end{table}

~\autoref{tab1:main} presents the results of Z1 and other models on 4 benchmarks, highlighting the following salient observations: 

\textbf{(1)} \textbf{Z1 models achieve comparable performance level with GPT-4o on  benchmarks of complex problems. (Avg. 45.4 vs 47.7)} This result highlights the success of test-time scaling with code, where performance improvements are achieved by leveraging extended reasoning traces during inference, rather than solely relying on increased model size.
\textbf{(2)} \textbf{Trained with trajectory data with code, Z1-7B outperforms other 7B-scale language reasoning models.} This outcome underscores the effectiveness of our test-time scaling approach, particularly when fine-tuned with code-realted reasoning data.
\textbf{(3)  Fine-tuning the model exclusively with code-related reasoning data  enables it to generalize across different domains.} Z1 models, fine-tuned on large amount of trajectories data with code,  displays superior generalization on GPQA Diamond (47.5\%) and MATH500 (76.4\%). This suggests the effectiveness of code-related trajectory training for language reasoning elicitation.

\begin{table}[t]
\centering
\small
\renewcommand{\arraystretch}{1.08}
\caption{\textbf{Z1-Code-Reasoning-107K Data Ablations.} We use a maximum
of around 4,096 thinking tokens for all scores in this table. The Random Sampling method does not alter the average trajectory length of the training samples. Both length-greedy sampling methods (longest and shortest) utilize the same number of training tokens (74M).}
\label{tab:ab}
\begin{tabular}{l|c|cc|cc}
\toprule
\textbf{Subset}& \textbf{Full} & \multicolumn{2}{c|}{\textbf{Random}} & \textbf{Longest} & \textbf{Shortest} \\

% \textbf{Sampling Strategy} & Oracle & Random & Random & Longest & Shortest \\
\midrule
\multicolumn{6}{c}{Training Dataset} \\
\midrule
\textbf{Dataset Size (Samples)} & 107K & 16K  & 64K & 33K & 90K \\
\textbf{Dataset Size (Tokens)} & 124M & 19M & 74M & 74M & 74M \\
\textbf{Mean Trajectory Length} & 1,159 & 1,157  & 1,156  & 2,216 & 807 \\
\midrule
\multicolumn{6}{c}{Evaluation} \\
\midrule
\textbf{GPQA Diamond} & 47.5  & 40.9  & 41.9    & 42.4   & 39.4   \\
\textit{Average Thinking Time}  & 2,470 & 1,797 & 2,241 & 2,695 & 1,979 \\
\midrule
\textbf{LiveCode Bench} & 35.3  & 32.2 & 34.1  & 32.7  & 34.1 \\

\textit{Average Thinking Time}  & 866 & 864 & 811 &927 &763 \\
\midrule
\textbf{MATH 500} & 76.4& 72.4  & 74.4  & 77.2  & 73.8 \\
\textit{Average Thinking Time} & 1,185 & 1,046 & 1,118 & 1,229 & 1,030 \\
\midrule
\textbf{AVG} & 53.1 & 48.5 & 50.1 & 50.8 & 49.1 \\
\textit{Average Thinking Time}  & 1,507 & 1,236 & 1,390 & 1,617 & 1,257 \\
\bottomrule
\end{tabular}
\end{table}

% \begin{table}[t]
% \centering
% \renewcommand{\arraystretch}{1.08}
% \caption{\textbf{Z1-Code-Reasoning-107K Data Ablations.} We use a maximum
% of around 4,096 thinking tokens for all scores in this table.}
% \label{tab:ab}
% \resizebox{\linewidth}{!}{%
% \begin{tabular}{l|c|cc|c|c}
% \toprule
% \textbf{Subset} & \textbf{Full} & \multicolumn{2}{c|}{\textbf{Random Sampling}} & \textbf{Longest}& \textbf{Shortest}\\
% \midrule
% \multicolumn{6}{c}{Training Dataset} \\
% \midrule
% \textbf{Dataset Size (Samples)} & 107K & 16K  & 64K & 33K & 90K \\
% \textbf{Dataset Size (Tokens)} & 124M & 19M & 74M & 74M & 74M \\
% \textbf{Mean Trajectory Length} & 1,159 & 1,157 (-2)  & 1,156 (-2) & 2,216 (+1057) & 807 (-352) \\
% \midrule
% \multicolumn{6}{c}{Evaluation} \\
% \midrule
% \textbf{GPQA Diamond} & 47.5  & 40.9 (-6.6) & 41.9 (-5.6)  & 42.4 (-5.1) & 39.4 (-8.1) \\
% \textit{Average Thinking Time}  & 2,470 & 1,797 (-673) & 2,241 (-229) & 2,695 (+225) & 1,979 (-491) \\
% \midrule
% \textbf{LiveCode Bench} & 35.3  & 32.2 (-3.1) & 34.1 (-1.2)  & 32.7 (-2.6)  & 34.1 (-1.2) \\
% \textit{Average Thinking Time}  & 866 & 864 (-2) & 811 (-55) & 927 (+61) & 763 (-103) \\
% \midrule
% \textbf{MATH 500} & 76.4 & 72.4 (-4.0)  & 74.4 (-2.0)  & 77.2 (+0.8)  & 73.8 (-2.6) \\
% \textit{Average Thinking Time} & 1,185 & 1,046 (-139) & 1,118 (-67) & 1,229 (+44) & 1,030 (-155) \\
% \midrule
% \textbf{AVG} & 53.1 & 48.5 (-4.6) & 50.1 (-3.0) & 50.8 (-2.3) & 49.1 (-4.0) \\
% \textit{Average Thinking Time}  & 1,507 & 1,236 (-271) & 1,390 (-117) & 1,617 (+110) & 1,257 (-250) \\
% \bottomrule
% \end{tabular}
% }
% \end{table}

\subsection{Data Ablations}
To further investigate the critical factors influencing effective reasoning elicitation in training data, we designed an ablation study with random sampling and greedy sampling  strategies (as shown in Algorithm \autoref{alg:greedy}) and obtain representative subsets with two key factors  (\textbf{Mean Trajectory Length} and \textbf{Training Sample Size}) that influences Z1's reasoning elicitation:
(1) Random Sampling: Samples are chosen randomly, serving as a baseline for comparison. We conduct random sampling with varying sample sizes (16K, 64K) to assess the impact of training samples on efficient test-time scaling.
(2) Longest Greedy Sampling: At each step, we select only the samples with the highest token counts, ensuring the subset contains the longest training examples in terms of reasoning traces. 
(3) Shortest Greedy Sampling : At each step, we select only the samples with the lowest token counts, maximizing the number of samples included in the subset while adhering to the token budget.

For all subsets, we calculate the Mean Trajectory Length (MTL) as follows:
\begin{equation}
\label{eq:data}
    \mathrm{MTL} = \frac{1}{n}\sum^{n}_{i=1}{\mathrm{Length}_i \mathrm{(tokens)}}
\end{equation}

Where $n$ denotes the number of training samples and $\mathrm{Length}_i$ represents the trajectory length (tokens) of the $i$-th training sample.
% ~\autoref{eq:data} describes two straightforward factors influencing the training datasets: \textbf{training samples size} \textbf{and} \textbf{ trajectory length}.
For all benchmarks, we calculated the Average Thinking Time (ATT) as follows:
\begin{equation}
    \mathrm{ATT} = \frac{1}{n}\sum^{n}_{i=1}{\mathrm{Length}_i} \mathrm{(tokens)}
\end{equation}
Where $n$ denotes the number of problems in benchmarks and $\mathrm{Length}_i$ represents the trajectory length of the $i$-th problem.
We fine-tune Qwen2.5-Coder-7B-Instruct on these representative subsets.
~\autoref{tab:ab}  presents the evaluation results of models fine-tuned on different subsets of the training data, highlighting the following observations:
\paragraph{Impact of Mean Trajectory Length}
Under the same training budget of 74M tokens, we sample two subsets with different mean trajectory length using two strategies: longest-greedy and shortest-greedy sampling. As shown in ~\autoref{tab:ab}, the subset sampled via  longest-greedy strategy exhibits a significantly higher MTL (2,216) compared to the shortest-greedy subset (807). This difference in training trajectory length translates into notable performance distinctions during evaluation. Specifically, the model fine-tuned on the longest-greedy subset demonstrates a longer Average Thinking Time (AVG score: 1,617 vs. 1,257) and a higher Benchmark Score (AVG score: 50.8 vs. 49.1) compared to the model trained on the shortest-greedy subset.
These results underscore the critical role of Trajectory Length in the training set, suggesting that longer trajectories enhance the model’s capacity for test-time scaling by encouraging more deliberate and extended reasoning during inference.
\paragraph{Impact of Training Sample Size}
To investigate the effect of training sample size, we randomly sample subsets of varying sizes (16K and 64K) from the original 107K dataset and compare the resulting models performance. 
As shown in ~\autoref{tab:ab},  the model fine-tuned on the full 107K dataset achieves an Average Thinking Time of 1,507 and a Benchmark Score of 53.1, outperforming the model trained on the Random-64K subset (1,390 and 50.1, respectively). In contrast, the smallest subset, Random-16K, yields the shortest Average Thinking Time (1,236) and the lowest performance (48.5). This observation indicates that a larger training sample size increase the effective thinking time thereby enhancing its overall performance, despite their Mean Trajectory Length remaining nearly identical across the randomly sampled subsets (1,157 for 16K, 1,156 for 64K) and the full dataset (1,159). 
% This observation implies that larger datasets increase the effective thinking time , enhancing performance despite similar Mean Trajectory Lengths.

\begin{algorithm}[h]
\caption{Greedy Sampling (Longest or Shortest)}
\label{alg:greedy}
\begin{algorithmic}[1]
\State \textbf{Input:} Set of training samples $S$, token budget $B$, strategy $\text{mode} \in \{\text{``longest''}, \text{``shortest''}\}$
\State \textbf{Output:} Subset $S' \subseteq S$ based on selected strategy
\State Initialize $S' \gets \emptyset$, total tokens $T \gets 0$
\While{$S \neq \emptyset$ \textbf{and} $T < B$}
    \If{$\text{mode} = \text{``longest''}$}
        \State Find $s^* \in S$ with maximum token count
    \ElsIf{$\text{mode} = \text{``shortest''}$}
        \State Find $s^* \in S$ with minimum token count
    \EndIf
    \If{$T + \text{token\_count}(s^*) \leq B$}
        \State $S' \gets S' \cup \{s^*\}$
        \State $T \gets T + \text{token\_count}(s^*)$
    \EndIf
    \State $S \gets S \setminus \{s^*\}$
\EndWhile
\State \textbf{return} $S'$
\end{algorithmic}
\end{algorithm}

\subsection{Test-time Scaling Comparison}
\label{sec:test-time-scaling}

\begin{figure}[t]
    \centering
    \includegraphics[width=\linewidth]{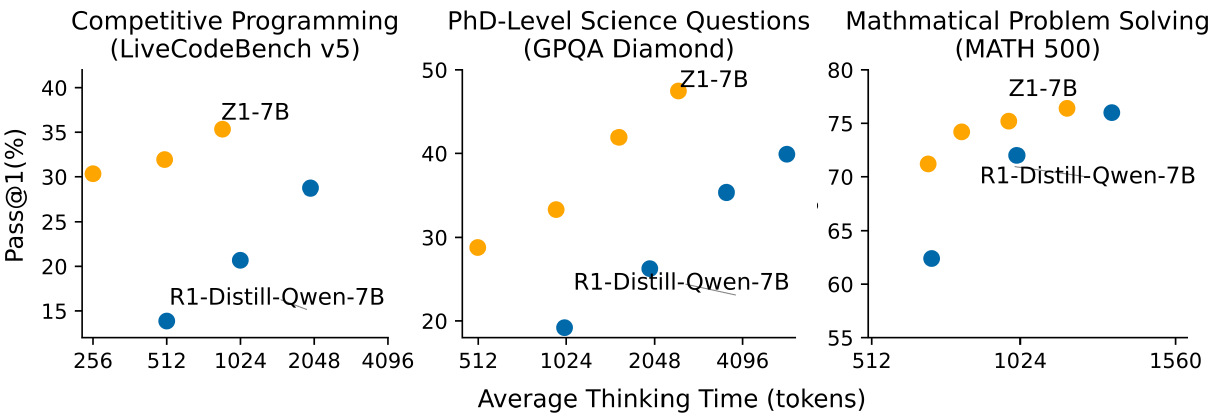}
    \caption{\textbf{Test-time scaling comparison between Z1-7B and R1-Distill-Qwen-7B.} R1-Distill-Qwen-7B can't adapt to the shifted thinking window without long and short trajectory fine-tuning. We budget force it by adding "the final answer is:".}
    \label{fig:comp-eff}
\end{figure}

To further compare the reasoning efficiency of Z1 to other models, we analyze the test-time scaling of Z1-7B and R1-Distill-Qwen-7B on three benchmarks: MATH500, GPQA, and LiveCodeBench. 
We equip Z1-7B with Shifted Thinking Window by imposing a cap of different maximum thinking tokens. 
For R1-Distill-Qwen-7B, We budget force~\cite{muennighoff2025s1} it by adding "the final answer is:", since R1-Distill-Qwen-7B can’t adapt to the shifted thinking window without long and short trajectory fine-tuning.
~\autoref{fig:comp-eff} illustrates the reasoning efficiency of Z1-7B compared to the baseline models, revealing the following key observations:

% \paragraph{Z1-7B outperforms Qwen2.5-Coder-7B-Instruct by leveraging code-related reasoning data for efficient test-time scaling. } 
% Compared to Qwen2.5-Coder-7B-Instruct, Z1-7B achieves consistently higher performance across all three reasoning task benchmarks by effectively utilizing increased thinking time. This improvement underscores the efficacy of code-related reasoning data in enabling test-time scaling. Notably, Z1-7B, trained solely on code-related reasoning trajectory data, demonstrates robust generalization to MATH500 and GPQA. This suggests that code-related trajectories possess significant potential for enhancing test-time scaling across diverse reasoning domains.

\paragraph{Z1-7B demonstrates more efficient test-time scaling than R1-Distill-Qwen-7B on reasoning tasks, by achieving comparable results with significantly fewer thinking tokens.}
Z1-7B exhibits superior test-time scaling efficiency compared to R1-Distill-Qwen-7B on reasoning tasks, by delivering better performance with significantly fewer thinking tokens. For example, Z1-7B outperforms R1-Distill-Qwen-7B while requiring only half the average thinking time (approximately 2,000+ tokens) on the GPQA Diamond benchmark, whereas R1-Distill-Qwen-7B relies on a budget exceeding 4,096 thinking tokens. This underscores Z1-7B’s remarkable efficiency in reasoning-intensive tasks. In contrast, R1-Distill-Qwen-7B demonstrates notably weaker performance at lower average thinking times (ATT), only achieving competitive results when ATT is substantially increased.

\subsection{Case Study}

\begin{figure*}[h]
\footnotesize
\centering

\begin{tabular}{@{}p{0.30\textwidth}p{0.30\textwidth}p{0.30\textwidth}@{}}
\toprule
\multicolumn{3}{@{}p{0.90\textwidth}@{}}{ % Spans all three columns
\textbf{Write a Python script to calculate the number of letter 'a' and 'r' in a string.}

Input: 

s = 'strawberry'

Output:

'a' : 1,
'r' : 3
} \\
\midrule % Optional: adds a line under the spanned header

\textbf{[Qwen2.5-Coder-7B-Ins]}

~

\textcolor{darkblue}{Here is a Python script that can solve the problem:}

\begin{Verbatim}[fontsize=\fontsize{7pt}{7pt}]
def count_letters(s):
    a_count = s.count('a')
    r_count = s.count('r')
    return {'a': a_count, 
            'r': r_count}

s = 'strawberry'
print(count_letters(s))
\end{Verbatim}

\textcolor{darkblue}{In this script, we define a function \texttt{count\_letters}...}

[120+ tokens omitted]

% Inside the function, we use the
% \texttt{count} method of the string to count the number of occurrences
% of `a' and `r'. We then return a dictionary with the counts of `a' and
% `r'.

% We then call this function with the string `strawberry' and print the
% result. The output will be:

% \begin{verbatim}
% {'a': 1, 'r': 3}
% \end{verbatim}

% This means that the string `strawberry' contains 1 `a' and 3 `r'.

&

\textbf{[Z1-7B]}

~

\textcolor{darkblue}{To solve this problem, we need...}
[50+ tokens omitted]

\textbf{1. Approach}

[94 tokens omitted]

\textbf{2. Solution Code}

\begin{Verbatim}[fontsize=\fontsize{7pt}{7pt}]
def count_letters(s):
    count_a = 0
    count_r = 0
    for char in s:
        if char == 'a':
            count_a += 1
        elif char == 'r':
            count_r += 1
    print("'a' :", count_a)
    print("'r' :", count_r)
\end{Verbatim}

\textbf{3.Explanation}

[104 tokens omitted]

\textcolor{darkblue}{This approach ensures that we...} 

[28 tokens omitted]

&

\textbf{[R1-Distill-Qwen-7B]}

~

\textcolor{red}{<think>}

\textcolor{red}{Okay, I need to write a Python script that counts the number of 'a's and 'r's...}

[1787 tokens omitted]

\textcolor{red}{</think>}

\textcolor{darkblue}{To solve this problem... }

[45 tokens omitted]

\textbf{1. Approach}

[92  tokens omitted]

\textbf{2. Solution Code}

\begin{Verbatim}[fontsize=\fontsize{8pt}{8pt}]
s = input().strip()
a_count = s.count('a')
r_count = s.count('r')
print(f'a: {a_count}')
print(f'r: {r_count}')
\end{Verbatim}

\textbf{3.Explanation}

[100+ tokens omitted]

\textcolor{darkblue}{This approach ensures that the solution is...}

[22 tokens omitted]

\\
\bottomrule
\end{tabular}

\vspace{5pt}

\caption{\textbf{Example model outputs. } We present a question that Qwen2.5-Coder-7B-Instruct (left) can correctly answer without requiring deep reasoning. The response from Z1 (middle) avoids excessive additional reasoning. In contrast, R1-Distilled-Qwen-7B generates 1,784 tokens of reasoning after the delimiter.}
% The gray ellipsis [...] indicates that the text was truncated to fit this page, though the full generated text is longer.}
\label{fig:ex}
\end{figure*}
~\autoref{fig:ex} illustrates a problem-solving example involving three models: Qwen2.5-Coder-7B, Z1-7B, and R1-Distill-Qwen-7B. 
In this instance, we present a simple problem, for which the Qwen2.5-Coder-7B-Instruct model delivers a correct answer accurately.
By comparison, R1-Distill-Qwen-7B adopts a context split to enforce thinking, requiring extensive deliberation that consumes 1,784 tokens before arriving at a solution. This protracted process underscores its inefficiency in optimizing thinking time for simpler tasks.
Z1-7B employs the Shifted Thinking Window to effectively bypassing unnecessary overthinking within the given context, demonstrating its advantage for balancing accuracy and efficiency in problem-solving.

% \subsection{Test-time Scaling Methods}

\section{Related Work}
\paragraph{Large Reasoning Models}
OpenAI o1 and o3 series models ~\cite{jaech2024openai}, which get trained with large-scale RL and learn to reason using chain-of-thought~\cite{o1}, have demonstrated strong reasoning ability in various complex downstream tasks with consistent gains from scaling test-time compute.
After the release of o1, Deepseek-R1~\cite{guo2025deepseek} replicates the performance of o1 through interleaved supervised fine-tuning and reinforcement learning. The R1-Distill series models, fine-tuned on samples distilled from DeepSeek-R1, also achieve test-time scaling through non-reinforcement learning (non-RL) approaches. In the realm of non-RL data distillation training, many open-source work, such as Sky-T1~\cite{sky_t1_2025}, s1~\cite{muennighoff2025s1}, and LIMO~\cite{ye2025limo} have successfully developed competitive reasoning models comparable to o1-preview. For reinforcement learning researches, models like QwQ-32B~\cite{qwq32b}, Kimi-K1.5~\cite{team2025kimik15}, and PRIME-7B~\cite{cui2025prime} have matched or even surpassed o1-preview’s performance. Our model, Z1, fine-tuned on 107K short and long distilled trajectory samples, employs the Shifted Thinking Window to enable weak reasoning for simple problems and strong reasoning for complex ones. Combining short and long trajectory training data and shifted thinking windows, our approach mitigates overthinking and achieves efficient test-time scaling.

\paragraph{Large Language Models for Code}
The development of large language models (LLMs) for code has undergone significant evolution from pre-trained models such as Codex, StarCoder~\cite{li2023starcoder}, DeepSeek-Coder~\cite{deepseek-coder}, and Qwen2.5-Coder Base~\cite{hui2024qwen25coder} to instruction-tuned variants like WizardCoder~\cite{luo2023wizardcoder}, WaveCoder~\cite{yu2024wavecoder}, and Qwen2.5Coder-Instruct~\cite{hui2024qwen25coder}. 
This evolution, with the advancement of test-time scaling, has led to a divergence in model capabilities.
On one hand, large reasoning models (LRMs) tailored for competitive programming (e.g., o1-Pro and o1-IOI~\cite{el2025competitive}) have emerged, leveraging chain-of-thought (CoT) reasoning to achieve human-level performance in programming contests. On the other hand, LLMs designed for software engineering (SE) tasks, such as Llama3-SWE-RL~\cite{wei2025swerl}, have been developed to address benchmarks like SWE-Bench~\cite{jimenez2023swe} and SWE-Lancer~\cite{miserendino2025swelancer}. These software engineering-focused LRMs incorporate real-world SE workflows (e.g., Agentless~\cite{xia2024agentless}) and reinforcement learning, progressively enabling automated project management.
In this work, we demonstrate that efficient test-time scaling with code can mitigate the tendency of LRMs to overthink coding problems. By optimizing thinking and answering strategies, we enhance model performance while reducing computational overhead, offering a novel perspective on the future directions of LLMs for code. Our findings contribute to both competitive programming and software engineering applications, bridging the gap between theoretical advancements and practical deployment.

\vspace{-0.1cm}
\section{Conclusion}
\vspace{-0.1cm}
In this work, we introduce an efficient test-time scaling method to elicit model reasoning abilities use fewer thinking tokens consumption. 
We train our Z1 model with a long and short code-related trajectory dataset and equip Z1 with shifted thinking window, a new approach to enable LRM to perform weak reasoning to simple problems and strong reasoning to complex problems.  
Trained with long and short trajectories and reasoning with shifted thinking window, Z1 matches state-of-the-art performance with comparable parameters and demonstrates efficient test-time compute scaling on various reasoning benchmarks. 
Furthermore, our systematic analysis of key factors for efficient reasoning elicitation provides valuable insights for future research, contributing to the development of more advanced and open-sourced reasoning models.

\newpage
\bibliographystyle{unsrt}
\bibliography{ref}
%%%%%%%%%%%%%%%%%%%%%%%%%%%%%%%%%%%%%%%%%%%%%%%%%%%%%%%%%%%%

\appendix
\newpage
\addtocontents{toc}{\protect\setcounter{tocdepth}{3}}
\renewcommand{\contentsname}{\Large Appendix Contents}
\tableofcontents
\clearpage

% \input{appendix-qwen-base}
% \clearpage
\section{Dataset Details}

\subsection{Comparison of Datasets}
We analyzed several recent open-source reasoning datasets. Table \ref{tab:datasets} presents several key characteristics of these datasets, including the number of samples, minimum and maximum token counts, domain, and dataset source. The Z1-Code-Reasoning dataset features shorter reasoning trajectories, which effectively enables our Z1 model to think quickly on simple problems while go into deeper reasoning on more challenging ones. This approach prevents overthinking and makes a significant contribution to achieving efficient test-time scaling.

\begin{table}[!h]
\small
\centering
\caption{The list of existing open-source reasoning datasets.}
\resizebox{\linewidth}{!}{
\begin{tabular}{cccccc}
\toprule
Dataset   & Samples &Min. tokens & Max. tokens& Domain & Dataset Source                        \\
\midrule
s1~\cite{muennighoff2025s1} & 1K & 667 & 7,850 & General & {\makecell{{Gemini 2.0 Flash} \\ {Thinking Experimental}}}    \\ \midrule
s1.1~\cite{muennighoff2025s1} & 1K & 923 & 26,685 & General & Deepseek R1  \\ \midrule
CodeForces-CoTs~\cite{penedo2025codeforces} & 48K & 523 & 25,156 & Competition Code & Deepseek R1  \\ \midrule
OpenR1-Math-220k\footnote{} & 22K & 4,307 & 18,611 & Math & Deepseek R1  \\ \midrule
OpenThoughts~\cite{openthoughts} & 114K & 299 & 91,198 & General & Deepseek R1  \\ 

\midrule
Z1-Code-Reasoning & 107K & 25 & 8,169 & General Code & QwQ-32B-Preview \\ 
\bottomrule
\end{tabular}
}
\label{tab:datasets}
% \vspace{-3ex}
\end{table}

\footnotetext[4]{\url{https://huggingface.co/datasets/open-r1/OpenR1-Math-220k}}

\subsection{Word frequency details}
In Section \ref{sec:data} and Figure \ref{fig:word_freq}, we analyzed the word frequency statistics of our dataset, Z1-Code-Reasoning-107k, in comparison to the previous Openthought-114k dataset. We performed a statistical analysis of word frequencies at the triplet level for both datasets, with the specific top 50 word frequencies detailed in Tables \ref{tab:word_freq1} and \ref{tab:word_freq2}. Due to the inclusion of reasoning data with varying trajectory lengths in our Z1-Code-Reasoning-107k dataset, there is a noticeable reduction in the overall word count. Additionally, as introduced in Section \ref{sec:data}, our dataset not only contains common logical reasoning connectives but also incorporates a greater proportion of code-related content.

\begin{longtable}[h]{lrlr}
\caption{Word frequency in Z1-Code-Reasoning-107K} 
\label{tab:word_freq1} \\
\toprule
trigram words & count & trigram words & count \\
\midrule
\endfirsthead
\caption[]{word frequency in 107K} \\
\toprule
trigram words & count & trigram words & count \\
\midrule
\endhead
\midrule
\multicolumn{4}{r}{Continued on next page} \\
\midrule
\endfoot
\bottomrule
\endlastfoot
i need to & 72090 & to ensure that & 18080 \\
we need to & 70123 & can use the & 17828 \\
the number of & 53932 & - for each & 15033 \\
\#\#\# explanation 1. & 37352 & code ```python def & 14631 \\
\#\#\# solution code & 33629 & a function that & 14000 \\
based on the & 31656 & iterate through the & 13698 \\
\#\#\# approach 1. & 30991 & you need to & 13485 \\
this approach ensures & 30213 & be able to & 13358 \\
to create a & 30150 & need to find & 13292 \\
approach ensures that & 28633 & how you can & 13280 \\
the sum of & 28373 & final solution to & 13149 \\
a list of & 28242 & to check if & 13119 \\
to solve this & 27732 & \#\# final solution & 13089 \\
solve this problem, & 26417 & \#\#\# explanation: 1. & 12451 \\
this problem, we & 26398 & keep track of & 12198 \\
solution code ```python & 25448 & sum of the & 12010 \\
problem, we need & 25446 & to calculate the & 11831 \\
to find the & 23880 & make sure that & 11800 \\
ensure that the & 22804 & the length of & 11610 \\
check if the & 21960 & we can use & 11448 \\
ensures that the & 21685 & need to create & 11366 \\
for i in & 20988 & to get the & 11150 \\
you can use & 20674 & need to handle & 11089 \\
the function should & 20524 & if there are & 11045 \\
- if the & 20434 & - we use & 11042 \\
\end{longtable}

\begin{longtable}[h]{lrlr}
\caption{Word frequency in OpenThoughts-114k} \label{tab:word_freq2} \\
\toprule
trigram words & count & trigram words & count \\
\midrule
\endfirsthead
\caption[]{word frequency} \\
\toprule
trigram words & count & trigram words & count \\
\midrule
\endhead
\midrule
\multicolumn{4}{r}{Continued on next page} \\
\midrule
\endfoot
\bottomrule
\endlastfoot
the number of & 603860 & the total number & 106781 \\
we need to & 484418 & equal to the & 106282 \\
the sum of & 321257 & which is the & 104615 \\
i need to & 188694 & the problem says & 98058 \\
in terms of & 173819 & at least one & 97784 \\
to find the & 169782 & let me try & 97748 \\
but the problem & 169082 & the distance from & 97535 \\
let me check & 155703 & but this is & 94306 \\
the answer is & 155213 & a + b & 91141 \\
need to find & 151935 & is less than & 90788 \\
for example, if & 151677 & so the code & 89894 \\
the problem is & 146089 & return your final & 89123 \\
is equal to & 142341 & your final response & 89120 \\
in this case, & 139293 & final response within & 89120 \\
\textbackslash) and \textbackslash( & 134985 & but let me & 87315 \\
sum of the & 133514 & both sides by & 86006 \\
according to the & 123939 & this is a & 85909 \\
but how to & 119825 & wait, but the & 85052 \\
if we can & 116574 & but since the & 83803 \\
but in the & 113484 & must be a & 81221 \\
but according to & 113061 & the midpoint of & 81158 \\
+ b + & 111663 & the same as & 80311 \\
a and b & 110523 & such that the & 80211 \\
the problem states & 107778 & with respect to & 78634 \\
total number of & 106790 & the angle between & 77260 \\
\end{longtable}

\clearpage
\section{Evaluation Details}
\subsection{Prompts for Evaluation}
We use prompts from Qwen2.5-Coder Github for LiveCodeBench~\footnote{\url{https://github.com/QwenLM/Qwen2.5-Coder/blob/main/qwencoder-eval/instruct/livecode_bench/lcb_runner/prompts/code_generation.py}}
and BigCodeBench~\footnote{\url{https://github.com/QwenLM/Qwen2.5-Coder/blob/main/qwencoder-eval/instruct/BigCodeBench/model.py}}
and the following prompts for MATH500 and GPQA Diamond:

\begin{tcolorbox}[enhanced jigsaw, breakable,
    colback=white, colframe=blue!20, coltitle=black,
    fonttitle=\small\bfseries,        % 标题字体调小
    % fontupper=\tiny,   % 正文字体设置为 scriptsize 等宽
    rounded corners,
    title=Prompts,
    listing engine=listings,
    listing options={language=Python, breaklines=true, breakatwhitespace=true, columns=flexible}]
% \begin{lstlisting}

<|im\_start|>system

Please reason step by step, and put your final answer within {\textbackslash}boxed\{\}.

<|im\_end|>

<|im\_start|>user

\{promblem\}

<|im\_end|>

<|im\_start|>assistant
% \end{lstlisting}
\end{tcolorbox}

% \begin{table}[ht]
%     \centering
%     \begin{tabular}{c|cc|cc|cc}
%     \hline
%         R1-7B & \multicolumn{2}{c|}{GPQA} & \multicolumn{2}{c|}{LiveCodeBench} & \multicolumn{2}{c}{MATH 500}  \\ \hline
%         max\_step & avg\_token & pass & avg\_token & pass & avg\_token & pass  \\ \hline
%         512 & 512.0 & 18.18 & 512.0 & 13.86 & 511.7 & 59.80  \\ 
%         720 & - & - & - & - & 719.2 & 62.40  \\ 
%         1024 & 1010.8 & 19.19 & 1022.5 & 20.68 & 1015.6 & 72.00  \\ 
%         1400 & - & - & - & - & 1339.6 & 76.00  \\ 
%         2048 & 1971.0 & 26.26 & 1977.7 & 28.75 & 1760.2 & 83.40  \\ 
%         4096 & 3606.7 & 35.35 & 3580.2 & 40.00 & 2540.8 & 87.60 \\ \hline
%     \end{tabular}
% \end{table}

% \begin{table}[ht]
%     \centering
%     \begin{tabular}{c|cc|cc|cc}
%     \hline
%         Z1-7B & \multicolumn{2}{c|}{GPQA} & \multicolumn{2}{c|}{LiveCodeBench} & \multicolumn{2}{c}{MATH 500}  \\ \hline
%         max\_step & avg\_token & pass & avg\_token & pass & avg\_token & pass  \\ \hline
%         256 & 256 & 25.25 & 256 & 30.34 & 255.12 & 37.2  \\ 
%         512 & 510 & 28.79 & 502 & 31.93 & 471 & 55.8  \\ 
%         1024 & 946 & 33.33 & 732 & 35 & 707 & 71.2  \\ 
%         2048 & 1548 & 41.92 & 865 & 35.34 & 1185 & 76.4  \\
%         ~ & 2470 & 47.47 & ~ & ~ & ~ & ~ \\ \hline
%     \end{tabular}
% \end{table}

\subsection{Test-time Values}
In Section \ref{sec:test-time-scaling}, we presented the results of Z1-7B and R1-Distill-Qwen-7B on three reasoning benchmarks, highlighting Z1-7B's more efficient test-time scaling capability. Table \ref{tab:evaluation} provides a more detailed account of the experimental results. By varying the maximum number of thinking tokens, we assessed the test-time scaling abilities of both models across different lengths of thinking trajectories. Z1-7B demonstrated strong performance even under tighter constraints on thinking tokens, showcasing its efficient test-time scaling capability. In contrast, R1-Distill-Qwen-7B required more thinking tokens to achieve a comparable level of performance.

\begin{table}[h]
    \centering
    \caption{Evaluation details on three reasoning benchmarks} 
    \label{tab:evaluation}
    \resizebox{\linewidth}{!}{
    \begin{tabular}{c|cc|cc|cc}
    \toprule
         \multirow{2}{*}{\textbf{Model}} & \multicolumn{2}{c|}{\textbf{GPQA}} & \multicolumn{2}{c|}{\textbf{LiveCodeBench}} & \multicolumn{2}{c}{\textbf{MATH 500}}  \\ 
        & \textit{Avg. think tokens} & \textit{Pass@1} & \textit{Avg. think tokens} & \textit{Pass@1} & \textit{Avg. think tokens} & \textit{Pass@1}  \\ \midrule
        \multirow{6}{1cm}{R1-Distill-Qwen-7B} & 512 & 18.2 & 512 & 13.9 & 512 & 59.8  \\ 
        & - & - & - & - & 719 & 62.4  \\ 
        & 1011 & 19.2 & 1023 & 20.7 & 1016 & 72.0  \\ 
        & - & - & - & - & 1337 & 76.0  \\ 
        & 1971 & 26.3 & 1978 & 28.8 & 1760 & 83.4  \\ 
        & 3607 & 35.4 & 3580 & 40.0 & 2541 & 87.6 \\ 
    \midrule
        \multirow{5}{*}{Z1-7B} & 256 & 25.3 & 256 & 30.3 & 255 & 37.2  \\ 
        & 510 & 28.8 & 502 & 31.9 & 471 & 55.8  \\ 
        & 946 & 33.3 & 732 & 35.0 & 707 & 71.2  \\ 
        & 1548 & 41.9 & 866 & 35.3 & 823 & 74.2  \\
        & 2470 & 47.5 & - & - & 1185 & 76.4 \\ 
    \bottomrule
    \end{tabular}
    }
\end{table}
\clearpage
\section{Examples}
\subsection{Examples on GPQA Diamond}
\begin{figure}[ht]
    \centering
    \includegraphics[width=\linewidth]{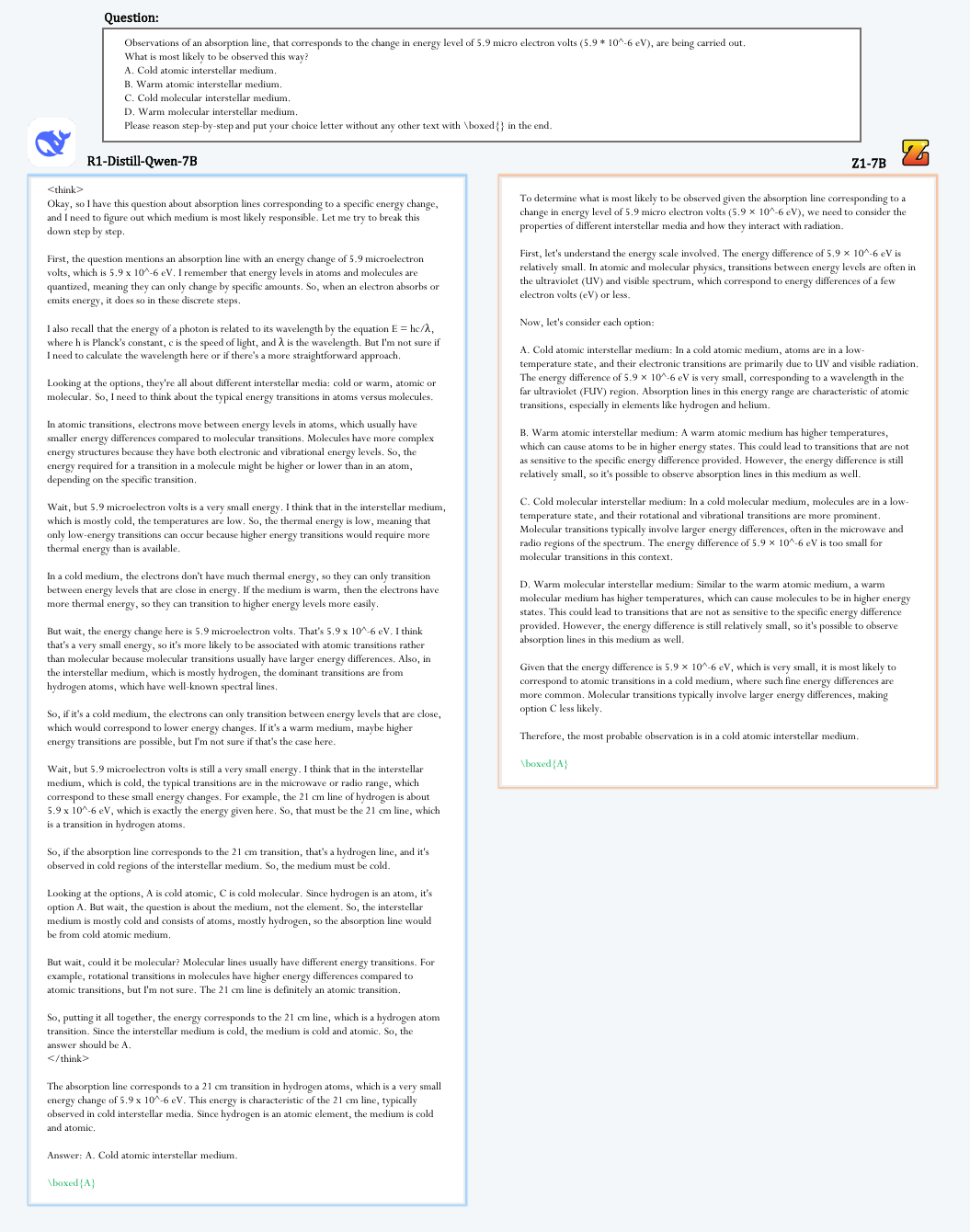} 
    \caption{Examples of the comparison between Z1-7B and R1-Distill-Qwen-7B on GPQA.}
    \label{fig:g1}
\end{figure}

\begin{figure}[ht]
    \centering
    \includegraphics[width=\linewidth]{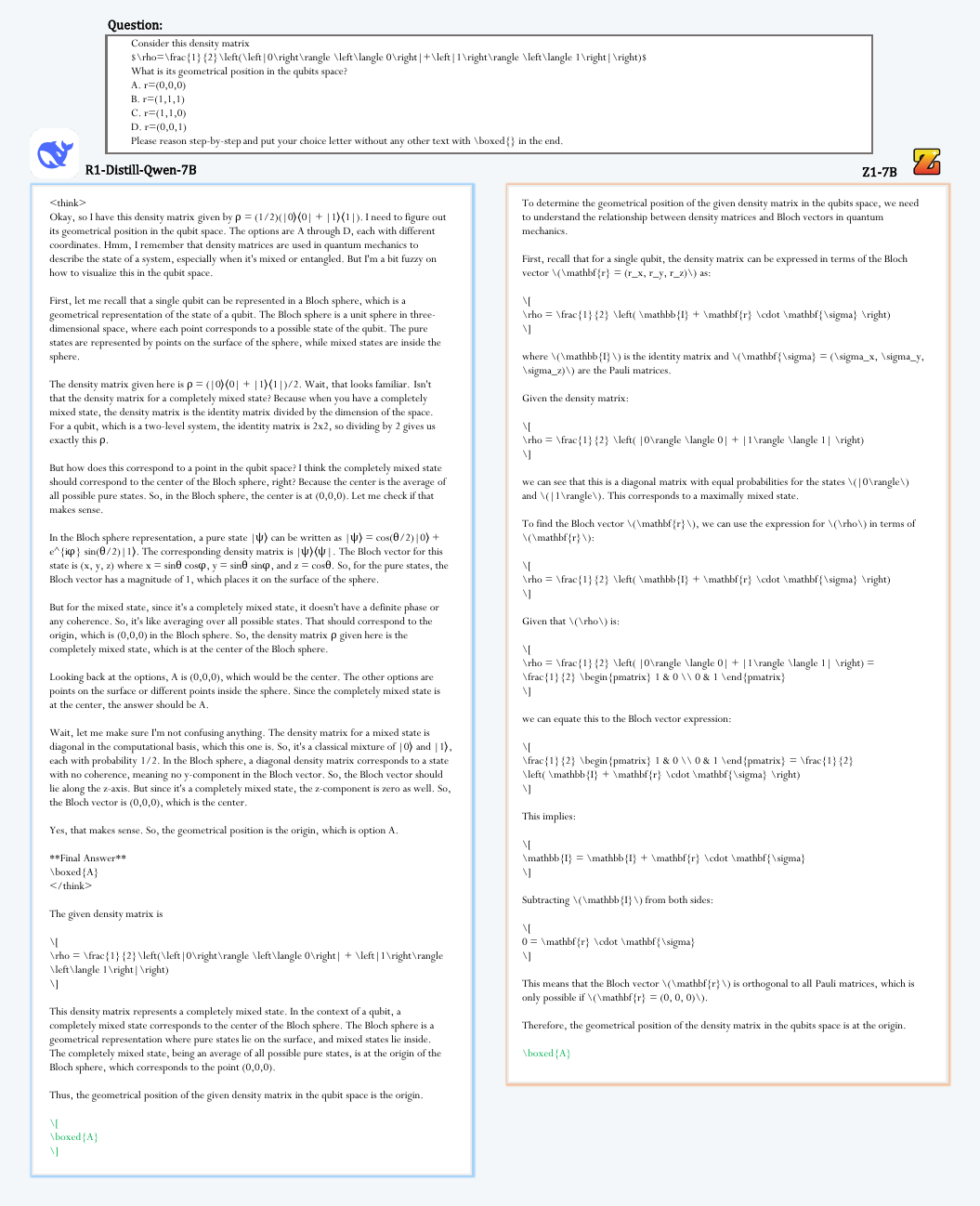} 
    \caption{Examples of the comparison between Z1-7B and R1-Distill-Qwen-7B on GPQA.}
    \label{fig:g2}
\end{figure}

\clearpage
\subsection{Examples on LiveCodeBench}
\begin{figure}[ht]
    \centering
    \includegraphics[width=\linewidth]{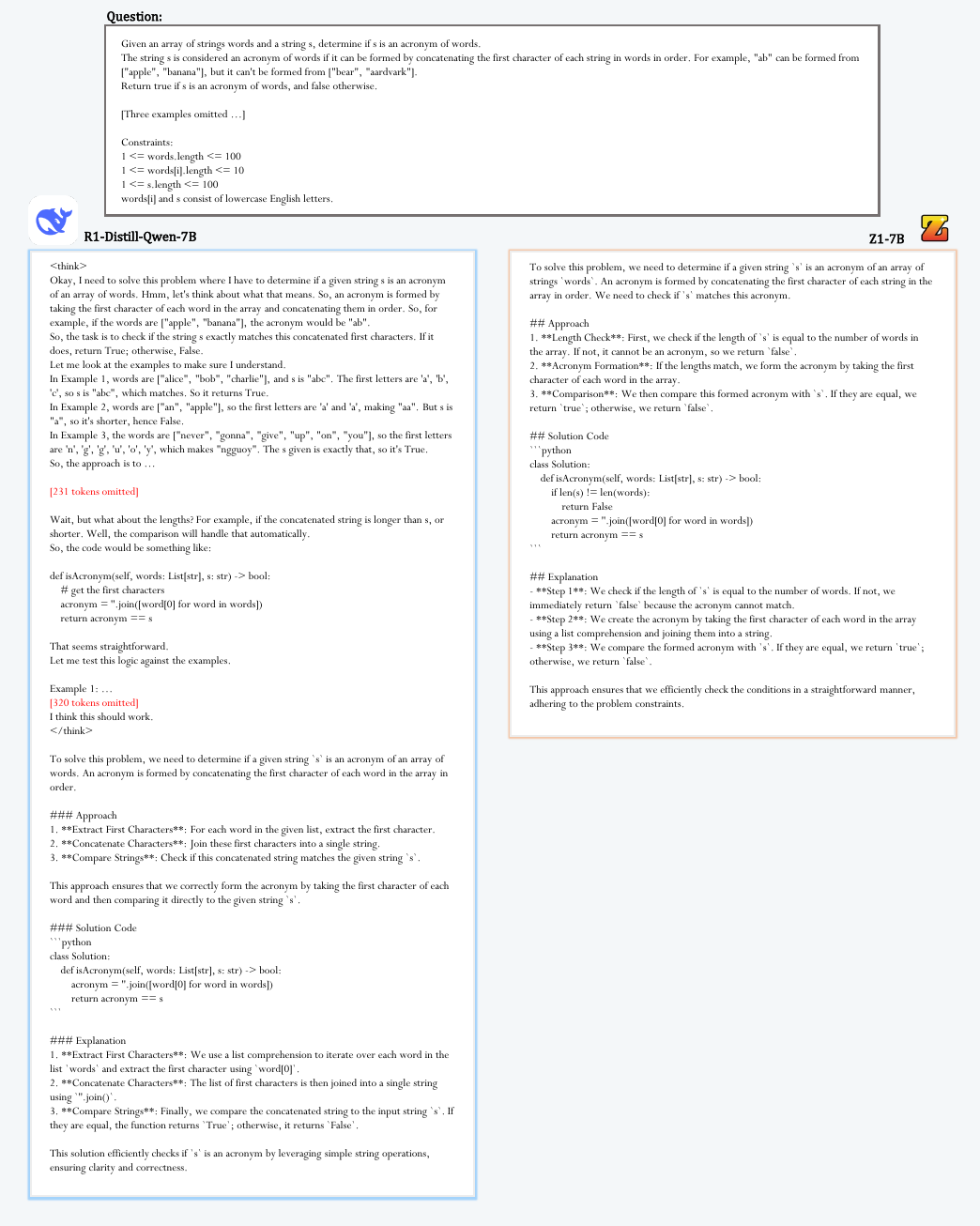} 
    \caption{Examples of the comparison between Z1-7B and R1-Distill-Qwen-7B on LiveCodeBench.}
    \label{fig:l1}
    
\end{figure}
\begin{figure}[ht]
    \centering
    \includegraphics[width=\linewidth]{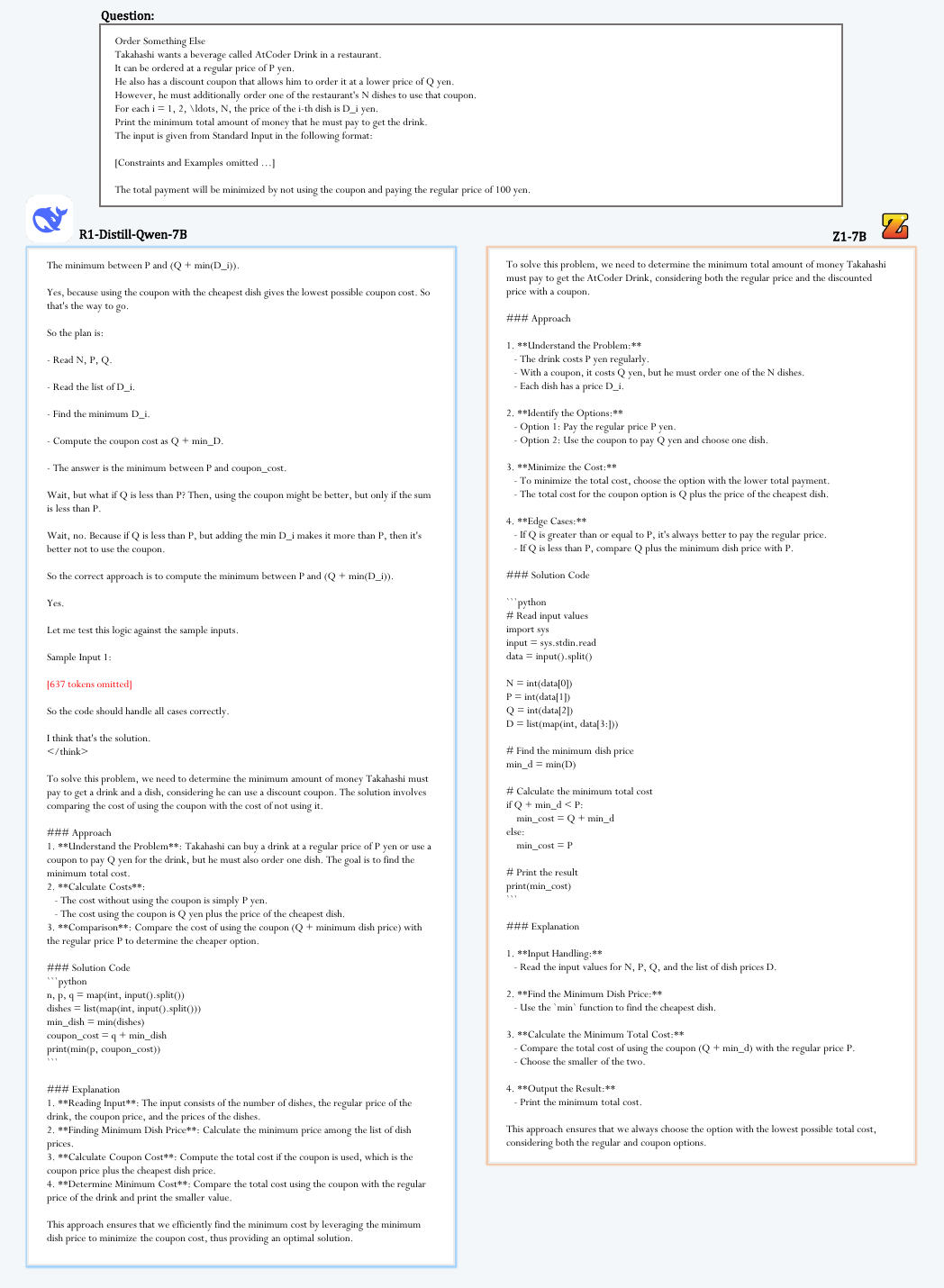} 
    \caption{Examples of the comparison between Z1-7B and R1-Distill-Qwen-7B on LiveCodeBench.}
    \label{fig:l2}
\end{figure}

\clearpage
\subsection{Examples on MATH500}
\begin{figure}[ht]
    \centering
    \includegraphics[width=\linewidth]{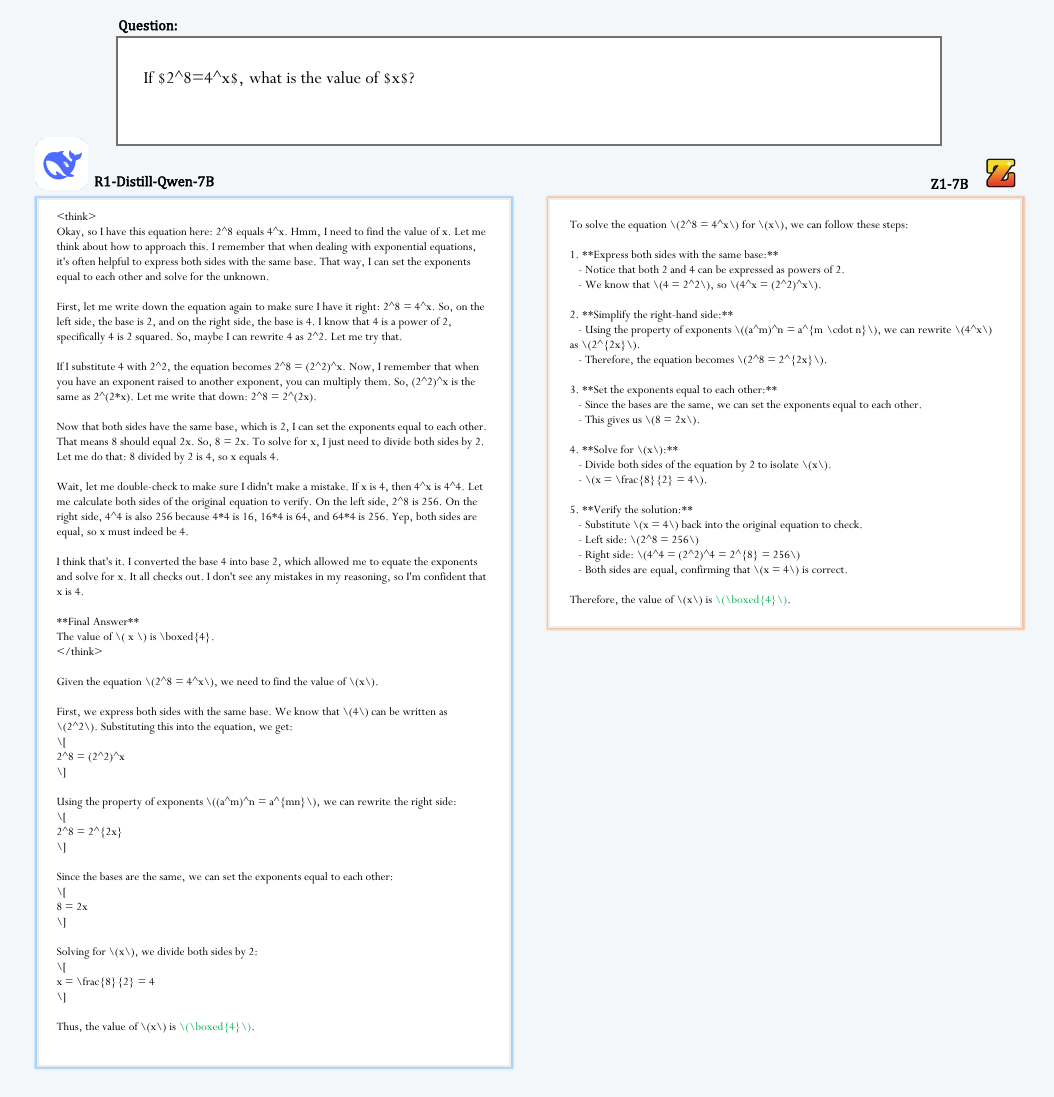} 
    \caption{Examples of the comparison between Z1-7B and R1-Distill-Qwen-7B on MATH500.}
    \label{fig:m1}
\end{figure}

\begin{figure}[ht]
    \centering
    \includegraphics[width=\linewidth]{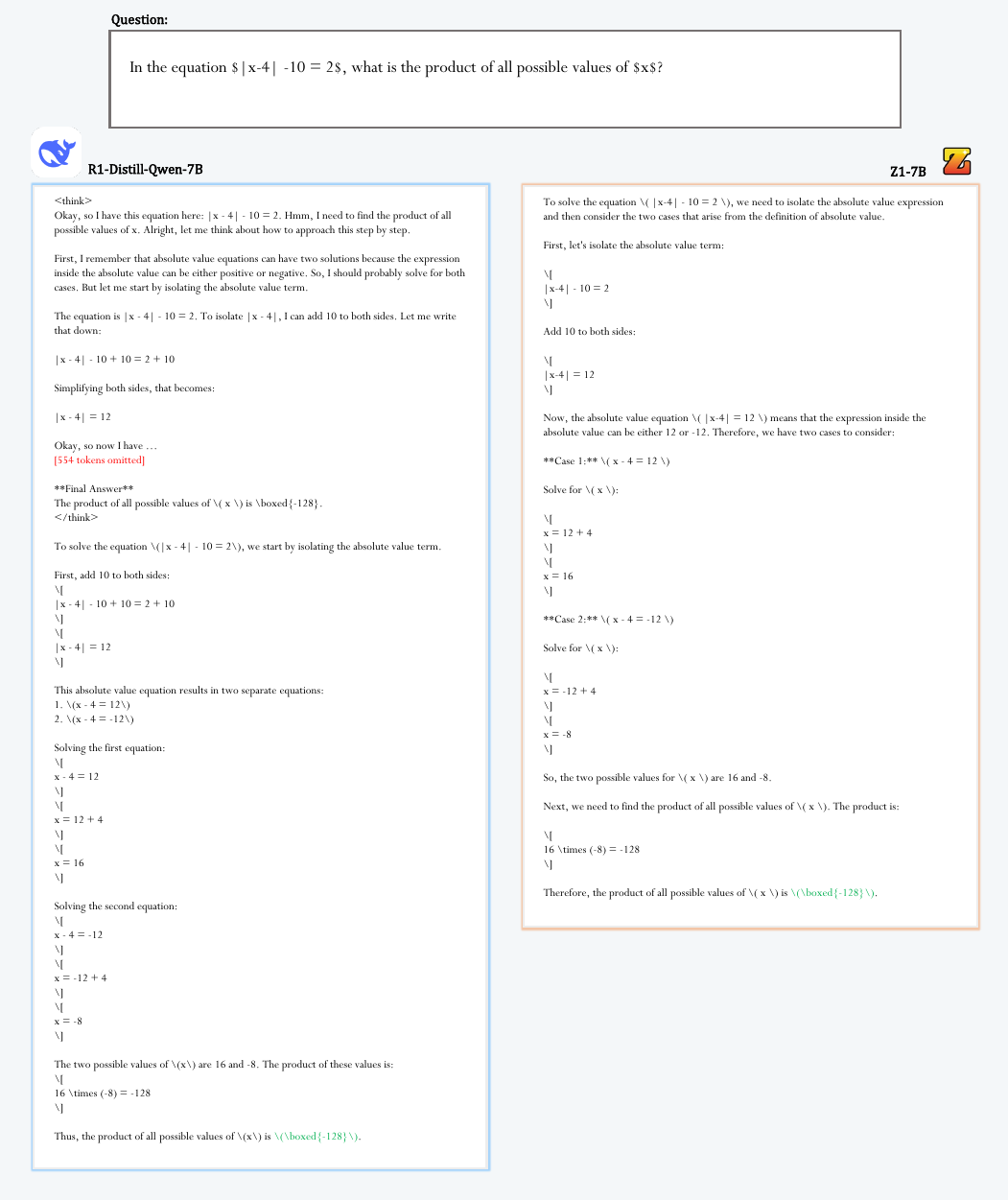} 
    \caption{Examples of the comparison between Z1-7B and R1-Distill-Qwen-7B on MATH500.}
    \label{fig:m2}
\end{figure}
% \newpage
% \input{appendix-case3}
%%%%%%%%%%%%%%%%%%%%%%%%%%%%%%%%%%%%%%%%%%%%%%%%%%%%%%%%%%%%

\end{document}